\documentclass{article}
\usepackage[preprint]{icml2026}

\usepackage{amsmath,amssymb,amsthm}
\usepackage{mathtools}
\usepackage{bm}
\usepackage{algorithm}
\usepackage{algorithmic}
\usepackage{booktabs}
\usepackage{multirow}
\usepackage{graphicx}
\usepackage{xcolor}
\usepackage{hyperref}
\usepackage{cleveref}
\usepackage{subcaption}
\usepackage{enumitem}
\usepackage{tikz}
\usepackage{comment}
\usetikzlibrary{arrows.meta,positioning,shapes.geometric,calc,fit,backgrounds}

\DeclareMathOperator*{\E}{\mathbb{E}}
\DeclareMathOperator{\KL}{KL}
\newcommand{\R}{\mathbb{R}}
\newcommand{\piref}{\pi_{\mathrm{ref}}}
\newcommand{\pitheta}{\pi_{\theta}}

\newcommand{\pigen}{\pi_{\mathrm{gen}}}
\newcommand{\loss}{\mathcal{L}}
\newcommand{\dataset}{\mathcal{D}}
\newcommand{\corpus}{\mathcal{C}}
\newcommand{\rubric}{\mathcal{R}}
\newcommand{\prompt}{\mathbf{x}}
\newcommand{\response}{\mathbf{y}}

\newcommand{\epsclip}{\epsilon_{\mathrm{clip}}}

\newtheorem{definition}{Definition}

\icmltitlerunning{Rubric-Grounded RL: Structured Judge Rewards for Generalizable Reasoning}

\begin{document}

\twocolumn[
\icmltitle{Rubric-Grounded Reinforcement Learning:\\
Structured Judge Rewards for Generalizable Reasoning in Language Models}

\icmlsetsymbol{equal}{*}

\begin{icmlauthorlist}
\icmlauthor{Manish Bhattarai}{t1}
\icmlauthor{Ismael Boureima}{t1}
\icmlauthor{Nishath Rajiv Ranasinghe}{t1}
\icmlauthor{Scott Pakin}{t1}
\icmlauthor{Dan O'Malley}{t1}
\end{icmlauthorlist}

\icmlaffiliation{t1}{Los Alamos National Laboratory, Los Alamos, NM, USA}

\icmlcorrespondingauthor{Manish Bhattarai}{ceodspspectrum@lanl.gov}

\icmlkeywords{Reinforcement Learning, GRPO, LLM-as-Judge, Reward Modeling, Generalization}

\vskip 0.3in
]

\printAffiliationsAndNotice{}

\begin{abstract}
We argue that decomposing reward into weighted, verifiable criteria and using an LLM judge to score them provides a partial-credit optimization signal: instead of a binary outcome or a single holistic score, each response is graded along multiple task-specific criteria. We formalize \emph{rubric-grounded reinforcement learning (RL)}: a framework in which the policy is optimized against a structured, multi-criterion reward produced by a frozen LLM judge that conditions on auxiliary grounding the policy never sees. We instantiate the framework by deriving rubrics from an Office of Scientific and Technical Information (OSTI)-derived corpus of roughly 100,000 scientific and technical documents and training Llama-3.1-8B-Instruct with Group Relative Policy Optimization (GRPO). With GRPO-based training, the model achieves $71.7\%$ normalized reward on held-out rubric evaluation. The GRPO-tuned policy also improves over the base model on four reasoning benchmarks not derived from the training corpus---GSM8K, MATH, GPQA Main, and GPQA Diamond. These results provide evidence that structured, document-grounded rewards can improve held-out rubric performance and induce transferable reasoning behaviors beyond the corpus used to construct the training environment.
\end{abstract}

\section{Introduction}
\label{sec:intro}

Reward design is the binding constraint in reinforcement learning of
large language models (LLMs). Standard alignment objectives often compress multi-faceted quality into
one scalar signal---a learned preference reward, a pairwise comparison, or
a binary verifier, and optimize the policy against that signal. This
works when the target behavior can be summarized by one latent utility,
but it under-specifies the optimization target whenever quality is
naturally decomposable: a strong technical answer must state the right
conclusion, use precise terminology, respect methodological caveats, and
connect evidence; a strong code review must catch correctness bugs, flag
style issues, and surface design concerns. This discarded structure could otherwise provide partial-credit learning
signals.

We study a general principle: \emph{decompose reward into weighted,
verifiable criteria, use an LLM judge to score them, and optimize the
policy with Group Relative Policy Optimization
(GRPO)~\citep{shao2024deepseekmath}}. We call this \emph{rubric-grounded
reinforcement learning}. The framework is domain-agnostic: any task whose
quality is plausibly written as a checklist of weighted criteria
(technical Q\&A, clinical summarization, legal drafting, pedagogical
assessment, structured code review) admits a rubric-grounded reward.

To validate the principle, we instantiate it on a concrete and scalable
data source: an OSTI-derived collection of roughly 100,000 scientific and
technical documents~\footnote{https://www.osti.gov/}. Each document is converted offline into
question--rubric pairs, where the rubric decomposes evaluation into
weighted criteria with required elements, scoring guides, and
verification cues. During training the policy answers each question
\emph{without} the source passage; a frozen judge scores each response
\emph{with} the passage and rubric. This information asymmetry encourages the policy to learn response
patterns that satisfy grounded criteria, rather than relying on access to
the source passage at rollout time.

\paragraph{Central claim.}
The empirical claim is that rubric-grounded GRPO can improve a
Llama-3.1-8B-Instruct model on held-out rubric tasks and yield gains on
external reasoning benchmarks. We support this with held-out rubric
evaluation, in-domain comparison to the supervised warm-start, four
reasoning benchmarks against the base model, and training/held-out reward
dynamics.

\paragraph{Contributions.}
\begin{enumerate}[leftmargin=*,topsep=2pt,itemsep=1pt]
  \item A \textbf{general framework}, rubric-grounded RL, that uses
        weighted criterion-level judge scores as a partial-credit reinforcement
        learning signal (\S\ref{sec:framework}).
  \item A \textbf{reward-to-RL mechanism} that converts passage-grounded
        criterion scores into group-relative GRPO advantages while
        preserving partial-credit structure
        (\S\ref{sec:reward}--\S\ref{sec:mechanism}).
  \item A \textbf{scalable instantiation}, document-derived rubrics, that
        produces RL training data without per-criterion human annotation
        (\S\ref{sec:instantiation}).
  \item An \textbf{empirical evaluation} showing that a
        rubric-grounded GRPO policy improves held-out rubric reward and
        yields gains on reasoning benchmarks not derived from the training
        corpus relative to the base instruction model
        (\S\ref{sec:experiments}--\S\ref{sec:analysis}).
\end{enumerate}

\begin{figure*}
    \centering
    \includegraphics[width=0.72\linewidth]{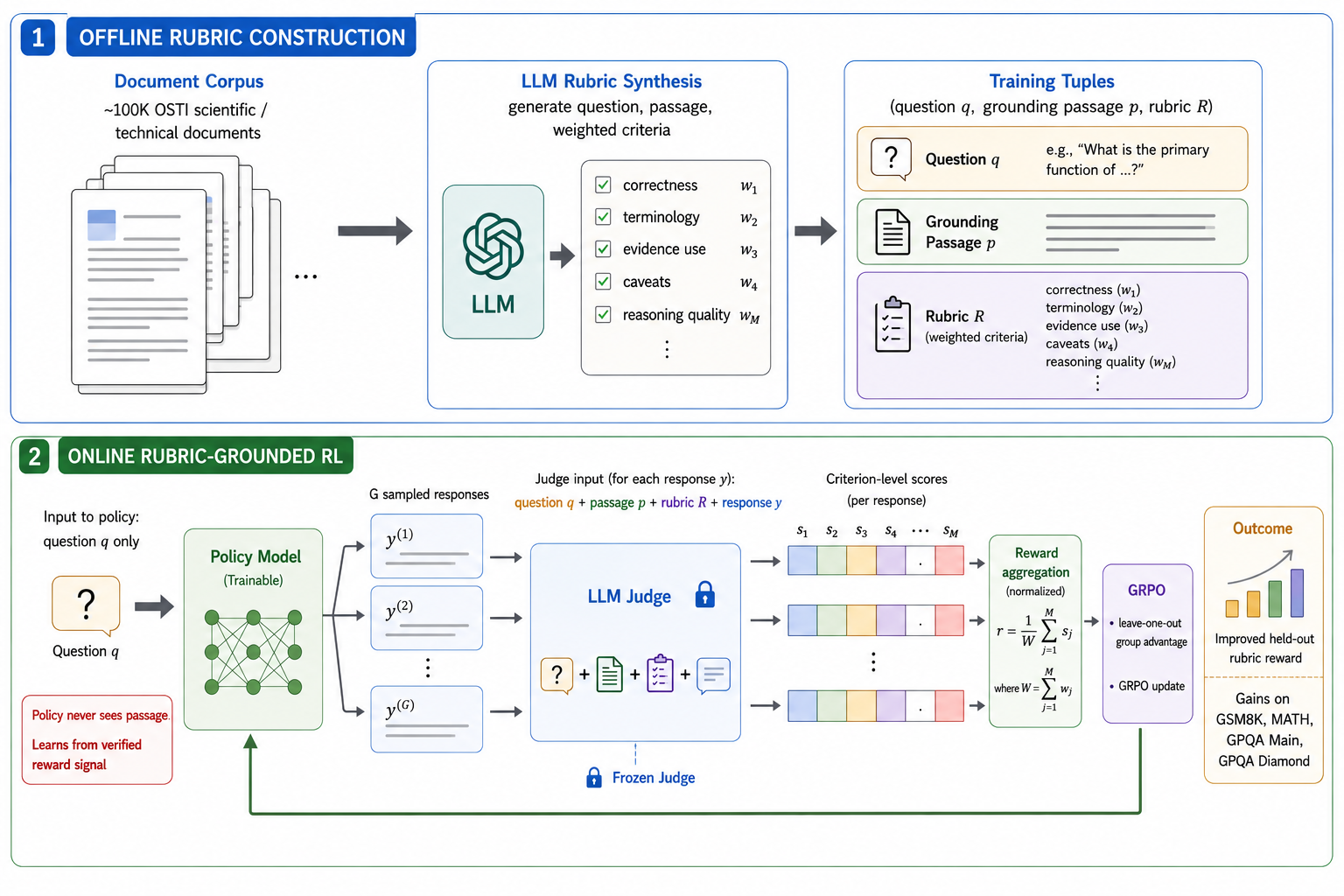}
    \caption{\textbf{Rubric-Grounded Reinforcement Learning pipeline.} Offline, a corpus of scientific and technical documents is used to synthesize training tuples consisting of a question, grounding passage, and weighted rubric. Online, the policy model receives only the question and generates multiple candidate responses. A frozen LLM judge evaluates each response using the hidden grounding passage and rubric, producing criterion-level scores that are aggregated into normalized rewards. These structured rewards provide the GRPO training signal, enabling the policy to learn from passage-grounded verification without accessing the passage at inference time.}
    \label{fig:overview}
\end{figure*}

\Cref{fig:overview} summarizes the framework end to end.

\section{Background and Related Work}
\label{sec:related}

\paragraph{Reward modeling for LLM alignment.}
Reinforcement Learning from Human Feedback (RLHF)~\citep{christiano2017deep,ouyang2022training,stiennon2020learning}
trains a scalar reward model on human pairwise preferences and optimizes
the policy with PPO~\citep{schulman2017proximal}. DPO~\citep{rafailov2023direct} and related preference-optimization methods
sidestep explicit reward modeling but still operate on pairwise
comparisons. Reinforcement Learning from AI Feedback (RLAIF)~\citep{bai2022constitutional,lee2023rlaif} replaces
human preferences with model-generated ones, retaining the scalar form.
Our framework departs from this scalar lineage: the reward is an
explicitly weighted vector of criterion scores, not a single learned
utility, and is tied to evaluation criteria specified at task-construction
time rather than inferred from preferences.

\paragraph{LLM-as-judge and rubric-based evaluation.}
LLM-as-judge methods have become a standard evaluation
device~\citep{zheng2023judging}, with recent work emphasizing structured
or rubric-based grading
\citep{kim2024prometheus,kim2024prometheus2,ye2024flask,cook2024checkeval}.
Most of this literature uses judges \emph{for evaluation only}. We use a
rubric-conditioned judge inside the optimization loop, which makes the judge
reliability and rubric design first-class methodological concerns rather
than evaluation details.

\paragraph{Rubrics as reinforcement-learning rewards.}
Recent work is closest to ours in using rubrics as RL reward signals:
Rubrics as Rewards extends Reinforcement Learning with Verifiable Rewards (RLVR) beyond automatically verifiable domains
with rubric-based feedback~\citep{gunjal2025rubrics}, Rubric Anchors
studies large-scale rubric construction and rubric-based RL for open-ended
tasks~\citep{huang2025rubricanchors}, and Rubric-Scaffolded RL uses
checklist rubrics both as rollout scaffolds and reward references
\citep{zhou2025rubricscaffolded}. Our setting differs in three ways: rubrics are synthesized from scientific
documents, the policy never sees the source passage during rollout, and
rubric synthesis is decoupled from the cheaper judge used repeatedly
during RL.

\paragraph{Process and step-level rewards.}
Process reward models (PRMs) provide dense per-step
supervision~\citep{lightman2023lets,wang2024mathshepherd,uesato2022solving} and have produced strong reasoning gains, but they typically require
human-annotated step labels or oracle verifiers and are usually scoped to
math. Rubric-grounded rewards generalize the dense-supervision idea
beyond per-step labels: criteria can be at any granularity (an entire
limitation discussion, a specific terminological requirement, a
verification cue) and require only a written rubric, not labeled
trajectories.

\paragraph{Group-relative policy optimization.}
GRPO~\citep{shao2024deepseekmath} replaces the learned value baseline
with a group-relative one computed from multiple rollouts per prompt,
which is a natural fit for judge-scored responses. Decoupled Clip and Dynamic sAmpling Policy Optimization (DAPO)~\citep{yu2025dapo}
and Group Sequence Policy Optimization (GSPO)~\citep{zheng2025gspo} refine the sampling and importance-sampling machinery. We use GRPO as the optimizer; the contribution lies in the
structure of the reward, not the optimizer.


\section{Method}
\label{sec:framework}

We present rubric-grounded RL in domain-agnostic terms; then instantiate  it with OSTI-derived documents. A task instance is a tuple $(\prompt, g, \rubric)$ where $\prompt$ is the
input shown to the policy, $g$ is auxiliary \emph{grounding} (a passage,
a reference solution, a code repository, a clinical chart) shown only to
the judge, and $\rubric$ is a structured rubric. The policy
$\pitheta(\response\mid\prompt)$ produces a response $\response$ without
access to $g$, while a frozen judge $\mathcal{J}$ scores $\response$ against
$\rubric$ using $g$ as grounding.

\begin{definition}[Rubric]
\label{def:rubric}
A rubric $\rubric=\{c_1,\ldots,c_M\}$ is a collection of criteria, each
$c_j=(w_j,\eta_j,\mathcal{E}_j,\kappa_j,\nu_j)$ with non-negative weight
$w_j\in\R_{\geq 0}$, natural-language description $\eta_j$, required
elements $\mathcal{E}_j$, expected keywords $\kappa_j$, and verification
method $\nu_j$. The total weight is $W=\sum_j w_j$.
\end{definition}

The policy and judge deliberately see different information: the policy
answers from $\prompt$ alone, while the judge sees
$(\prompt, g,\response,\rubric)$. This separates what the policy must
learn to produce from what the judge may verify. The same template applies
whenever quality can be written as weighted, locally checkable criteria:
technical question answering grounded in source documents, clinical
reasoning grounded in patient charts, code review grounded in a diff and
tests, legal drafting grounded in statutes and precedent, or pedagogical
feedback grounded in a marking scheme. The following subsections define
the reward, show how it enters GRPO, explain why the resulting scalar is
more informative than a flat holistic score, and describe the OSTI-derived
document instantiation.

\subsection{Reward Construction}
\label{sec:reward}

\paragraph{Multi-Criterion Judge Reward}

Given $(\prompt, g, \rubric)$ and a sampled response
$\response\sim\pitheta(\cdot\mid\prompt)$, the judge produces a vector of
per-criterion scores
\begin{equation}
  \label{eq:judge_scores}
  \{s_j\}_{j=1}^M = \mathcal{J}(\prompt, g, \response, \rubric),
  \quad s_j \in [0, w_j],
\end{equation}
where $s_j$ is the awarded weight on criterion $c_j$. The aggregate raw
reward is $r_{\mathrm{raw}}=\sum_j s_j$.

\paragraph{Normalization}

To compare rubrics with different total weights, we normalize:
\begin{equation}
  \label{eq:norm_reward}
  r(\prompt, \response) = \frac{1}{W}\sum_{j=1}^M s_j \;\in\; [0, 1].
\end{equation}
Parsed rewards are clipped to $[0,1]$ in implementation to handle schema
or numerical edge cases.

\paragraph{Judge Prompt Architecture}

The judge prompt contains a strict-evaluator system instruction, the
grounding $g$ (truncated to fit context), the policy input $\prompt$, the
response $\response$, the rubric serialized as a numbered list with each
criterion's weight, description, required elements, scoring guide,
keywords, and verification method, and a structured-output specification
requiring JSON with per-criterion scores, total, max, and a brief
justification. The judge is decoded at low temperature
($\tau_{\mathcal{J}}=0.1$) to reduce reward variance. Full prompt
templates are in Appendix~\ref{app:judge_prompt}.

Rubric synthesis and reward judging are decoupled. The rubrics can be
created offline with a stronger or more expensive model, while online RL
uses a cheaper judge repeatedly for reward computation. This separation is
important because rubric construction is paid once, whereas judging is
paid at every GRPO step.




Each GRPO step requires $B\times G$ judge calls, which dominate
wall-clock time. We deploy $N_{\mathrm{workers}}$ parallel judge actors,
each holding an independent client to a vLLM-style inference endpoint;
workers process micro-batches of size $B_{\mathrm{judge}}$ in parallel.
This separates policy training (GPU-bound, autoregressive) from judge
evaluation (latency-bound, remote).

\subsection{Rubric-Grounded Policy Optimization}
\label{sec:policy}

The optimizer receives only a scalar reward, but that scalar is produced
by a structured measurement process: the judge scores each sampled answer
against the same passage-grounded rubric and then aggregates criterion
scores. This section describes how that structured reward is converted
into GRPO updates in the reported experiments. Implementation
variants that were supported but not used in the main run are deferred to
Appendix~\ref{app:variants}.

\paragraph{Training Objective}

For policy $\pitheta$ and frozen reference $\piref$,
\begin{equation}
  \label{eq:objective}
  \max_{\theta}\;
  \E_{\prompt\sim\dataset,\,\response\sim\pitheta(\cdot\mid\prompt)}
  [r(\prompt,\response)]
  - \beta\,\E_{\prompt}[\KL(\pitheta\,\|\,\piref)],
\end{equation}
with $r$ from~\eqref{eq:norm_reward}.

\paragraph{Group-relative rubric credit.}

For each prompt $\prompt_i$, we sample $G$ responses
$\{\response_i^{(g)}\}_{g=1}^G\sim\pigen(\cdot\mid\prompt_i)$. Each
response is judged against the same grounding $g_i$ and rubric
$\rubric_i$, producing rewards $\{R_i^{(g)}\}$. Thus, the update compares
answer strategies under a fixed assessment context rather than comparing
raw scores across unrelated questions. The leave-one-out (LOO) baseline
and group standard deviation are
\begin{align}
  b_i^{(g)} &= \tfrac{1}{G-1}\sum_{g'\neq g} R_i^{(g')},\\
  \sigma_i &= \mathrm{std}\bigl(\{R_i^{(g')}\}_{g'=1}^G\bigr),
\end{align}
yielding the normalized advantage
\begin{equation}
  \label{eq:advantage}
  A_i^{(g)} =
  \begin{cases}
    \dfrac{R_i^{(g)}-b_i^{(g)}}{\sigma_i+\delta} & \sigma_i>0,\\[4pt]
    0 & \text{otherwise.}
  \end{cases}
\end{equation}
The advantage is broadcast to every response token.

\paragraph{Information-asymmetric update.}

The asymmetry from \S\ref{sec:framework} enters the update through
$R_i^{(g)}$: rewards are computed by a judge that sees
$(\prompt_i,g_i,\response_i^{(g)},\rubric_i)$, but gradients update a
policy that generated $\response_i^{(g)}$ from $\prompt_i$ alone. The
policy cannot retrieve the passage at inference time, but it is trained
toward responses that satisfy passage-grounded criteria during learning.
The result is a training-time supervision signal closer to an examiner
with an answer key than to a retrieval-augmented generator.

\paragraph{Clipped surrogate, KL, and stabilizers.}

With per-token ratio $r_t=\exp(\log\pitheta-\log\pigen)$, the clipped
surrogate is
\begin{equation}
  \label{eq:clipped_loss}
  \ell_t^{\mathrm{clip}} = -\min\!\bigl(r_t\,A_t,\,
  \mathrm{clip}(r_t, 1-\epsclip, 1+\epsclip)\,A_t\bigr),
\end{equation}
and the per-token KL uses the $k_3$ approximation~\citep{schulman2020kl}:
\begin{equation}
  \label{eq:kl_approx}
  \widehat{\KL}_t = e^{u_t}-1-u_t,\quad
  u_t = \log\piref - \log\pitheta.
\end{equation}
Inputs are clamped to $[-20,20]$ for numerical stability.

\paragraph{Total loss.}
Combining the clipped policy term and KL penalty gives the token-averaged
training loss
\begin{equation}
  \label{eq:total_loss}
  \boxed{
  \loss(\theta) = \frac{1}{\sum_{i,g}T_{ig}}\sum_{i,g,t} m_t^{(ig)}
  \!\left[\ell_t^{\mathrm{clip}}+\beta\,\widehat{\KL}_t\right],
  }
\end{equation}
where $m_t^{(ig)}$ masks padding and prompt tokens. The complete algorithm is given in Algorithm~\ref{alg:grpo} (Appendix~\ref{app:algorithm}).
In addition to KL regularization, the implementation uses low-temperature
judging, schema-constrained parsing, reward clipping, and conservative
zero reward on parse failures. These choices are not separate
contributions; they are safeguards that make the structured reward usable
inside an online RL loop.

\subsection{Mechanism: Structured Credit Assignment}
\label{sec:mechanism}

The central object is not the final scalar reward but the structured
measurement that produces it. For a rubric with positive-weight criteria
$w_1,\ldots,w_M$ and total weight $W=\sum_j w_j$, define normalized
criterion scores
\begin{equation}
  z_j(\prompt,g,\response)
  = \frac{s_j(\prompt,g,\response,\rubric)}{w_j}\in[0,1],
  \qquad
  \alpha_j=\frac{w_j}{W}.
\end{equation}
The reward used by GRPO is the projection
\begin{equation}
  \label{eq:rubric_projection}
  r(\prompt,\response)
  = \sum_{j=1}^M \alpha_j z_j(\prompt,g,\response)
  = \alpha^\top z(\prompt,g,\response).
\end{equation}
Thus rubric-grounded RL is scalar RL over a reward that is first
factorized into interpretable criterion coordinates. The mechanism below
spells out what this factorization buys.

\paragraph{Resolution: more ways to be partly right.}

Let each criterion score $z_j$ take values on a finite grid
$\mathcal{Z}_j$. A binary verifier exposes at most two reward levels. A
holistic judge exposes one scalar axis. A rubric exposes the product
space
\begin{equation}
  \mathcal{Z}_{\rubric}
  = \mathcal{Z}_1\times\cdots\times\mathcal{Z}_M
\end{equation}
before projection by~\eqref{eq:rubric_projection}. Even when different
criterion vectors collide after weighting, the judge has evaluated the
response through a higher-resolution measurement. When weighted sums do not collide, this yields up to
$\prod_j |\mathcal{Z}_j|$ attainable scalar reward levels rather than
two.

This is the mathematical version of partial credit: an answer can move
from $z=(0,0,0)$ to $(1,0,0)$ without satisfying all criteria. Early in
RL, when most generations are imperfect, such intermediate rewards
produce useful ordering among bad, partial, and nearly complete answers.

\paragraph{Localization: criterion failures affect the scalar advantage.}

The optimizer does not receive separate per-criterion gradients; it
receives the scalar $\alpha^\top z$. However, the scalar advantage
changes in proportion to the weighted criterion differences between
sampled responses:
\begin{equation}
  \label{eq:criterion_delta}
  r(\response_a)-r(\response_b)
  = \sum_{j=1}^M \alpha_j
    \bigl[z_j(\response_a)-z_j(\response_b)\bigr].
\end{equation}
A missing required element in criterion $j$ changes the scalar reward by
exactly its weighted contribution $\alpha_j\Delta z_j$. The localization
therefore lives in the measurement, not in a new optimizer: GRPO still
updates on a scalar advantage, but that advantage is assembled from
explicit axes of quality rather than from a single free-form preference
judgment.

\paragraph{Privileged grounding: a training-time verifier}

The reward in~\eqref{eq:rubric_projection} depends on $g$, but the policy
does not condition on $g$:
\begin{equation}
  \response\sim\pitheta(\cdot\mid\prompt),
  \qquad
  r=\mathcal{J}(\prompt,g,\response,\rubric).
\end{equation}
This is a learning-with-privileged-information setup: grounding is
available to the verifier at training time, not to the policy at rollout
or evaluation time. Consequently, the reward can penalize unsupported
claims and missing facts without converting the policy into a
retrieval-augmented system. Successful behavior must therefore come from the policy's learned
parameters rather than from copying the hidden passage.

\paragraph{Noise attenuation by effective criteria.}

Suppose criterion scores are observed with independent zero-mean judge
noise: $\tilde{z}_j=z_j+\xi_j$, $\E[\xi_j]=0$,
$\mathrm{Var}(\xi_j)=\tau^2$. The aggregate reward noise is
\begin{equation}
  \label{eq:effective_criteria}
  \begin{aligned}
  \mathrm{Var}(\alpha^\top \xi)
  &= \tau^2\sum_j\alpha_j^2
   = \frac{\tau^2}{M_{\mathrm{eff}}}, \\
  M_{\mathrm{eff}}
  &= \frac{1}{\sum_j\alpha_j^2}
   = \frac{W^2}{\sum_j w_j^2}.
  \end{aligned}
\end{equation}
For equally weighted criteria, $M_{\mathrm{eff}}=M$: independent
criterion noise is averaged across the rubric. This is not a guarantee
of judge correctness, systematic or correlated judge errors do not
average away, but it explains why decomposed scoring is preferable to a
single holistic call when criterion errors are at least partly
independent.

\paragraph{Behavioral transfer hypothesis.}

The expected transfer is behavioral rather than factual: criteria such as
precision, constraint satisfaction, stepwise justification, grounded
terminology, and calibrated uncertainty may benefit external tasks that
reward similar behaviors. This motivates evaluating on disjoint reasoning
benchmarks (\S\ref{sec:results}).

\subsection{Instantiation: Document-Derived Rubrics}
\label{sec:instantiation}

We instantiate the framework with rubrics derived from an OSTI-derived
corpus of roughly 100,000 scientific and technical documents. This corpus is
a convenient source because the documents are abundant, naturally
grounded (each rubric points to a specific passage), and require no
per-criterion human annotation.

\paragraph{Document-to-rubric pipeline.}

The pipeline operates in three LLM-prompted stages:

\paragraph{Stage 1: Semantic analysis.}
For each document $d$, an LLM produces a structured representation
$\mathbf{s}(d)$ capturing genre, primary contribution, central concepts,
technical depth, and dominant reasoning mode. This representation
conditions downstream synthesis.

\paragraph{Stage 2: Joint question--rubric synthesis.}
Conditioned on $(\mathbf{s}(d), d)$, the LLM produces $K_d$ tuples
$\{(q_k, p_k, \rubric_k)\}$ subject to four constraints: questions and
rubrics must be self-contained without the source; questions must target
deep understanding rather than surface recall; rubrics must include all
five components from Definition~\ref{def:rubric}; and weights must
reflect the criterion's contribution to overall quality.

\paragraph{Stage 3: Rubric enrichment.} 
A post-processing step augments criteria with structured cues, expected
terminology, core concepts, and verification patterns to improve
inter-rater consistency.

\paragraph{Quality assurance.}

Generated tuples pass structural checks before inclusion: minimum
criteria per rubric, minimum total weight, non-empty question, all
criterion fields present, non-negative weights. Examples failing any
check are discarded. Corpus-level statistics (criteria-per-example,
weight distributions, length histograms) are tracked for reporting.

\paragraph{Scalable generation.}

Documents are dispatched with bounded asynchronous concurrency; outputs
are written incrementally and a content-hash index supports idempotent
resume. Full data-pipeline configuration is in
Appendix~\ref{app:data_pipeline}.

\section{Experimental Setup}
\label{sec:experiments}

\subsection{Dataset}
We construct a rubric-grounded dataset of approximately 100K document-grounded questions from OSTI scientific and technical documents. Each question is paired with a grounding passage and a structured rubric containing 5–10 weighted criteria. We partition the dataset into 70\% training, 15\% validation, and 15\% test splits. The training split is used for RL reward computation, the validation split for model selection and hyperparameter tuning, and the test split for held-out evaluation.

\subsection{Model and Infrastructure}

The policy is Llama-3.1-8B-Instruct~\citep{dubey2024llama3} with $\piref$
initialized from the same checkpoint. Question-rubric dataset generation
uses GPT-OSS-120B. Judge inference is served by a GPT-OSS-20B endpoint.
Training is distributed with separate generation and judge-evaluation
workers; the implementation is built on
NeMo-RL~\footnote{https://github.com/NVIDIA/NeMo-RL}.

\subsection{Hyperparameters}

For the main GRPO run, each optimization step uses $B=64$ prompts and
samples $G=32$ generations per prompt, yielding an effective batch size
of $2048$. We restrict rollouts to a single turn, and enable both the
leave-one-out baseline and reward normalization. The loss uses clipping parameter $\epsclip=0.2$, advantage stabilizer
$\delta=10^{-8}$, KL coefficient $\beta=0.01$, and the $k_3$
KL approximation. We optimize with AdamW at learning rate
$\eta=3{\times}10^{-7}$, weight decay $0.01$, and maximum gradient norm
$1.0$, using $13$ warmup steps followed by a constant learning-rate
schedule. The judge is run with $N_{\mathrm{workers}}=32$, temperature
$\tau_{\mathcal{J}}=0.1$, a token budget of $16{,}000$, a maximum passage
length of $50{,}000$ characters, and worker batch size $4$. Variants not
enabled in the main configuration are reported in
Appendix~\ref{app:variants}.

\subsection{Comparison Models}
\label{sec:exp:baselines}

We compare the trained policy against its initialization and, for
in-domain rubric evaluation only, the supervised warm-start:
\begin{itemize}[leftmargin=*,topsep=1pt,itemsep=0pt]
  \item \textbf{Base:} Llama-3.1-8B-Instruct, no further training.
  \item \textbf{SFT:} supervised warm-start on question--answer data from
        the same document-derived corpus.
  \item \textbf{Ours:} the same model after one rubric-grounded GRPO run.
\end{itemize}

\subsection{Evaluation Benchmarks}

We report (i)~held-out rubric reward on a random split of the synthesized
tuples (in-domain), and (ii)~standard accuracy on four reasoning
benchmarks not derived from the training corpus: GSM8K~\citep{cobbe2021training},
MATH~\citep{hendrycks2021measuring},
GPQA Main, and GPQA Diamond~\citep{rein2023gpqa}. Using benchmarks not
derived from the corpus is what makes them a transfer test rather than a
recall test. Base and Ours are evaluated with the same prompt template,
decoding parameters, and scoring script. We report single-sample
accuracy. GSM8K and MATH use exact-match final-answer scoring after
answer extraction; GPQA Main and GPQA Diamond use the standard
multiple-choice accuracy protocol. The reported GRPO checkpoint is
selected by held-out rubric reward.

\section{Results and Analysis}
\label{sec:results}

\subsection{In-Domain Judge Measurements}

\begin{table}[!ht]
\centering
\caption{\textbf{In-domain held-out rubric evaluation.}
All rows are evaluated on the same held-out question--rubric test split using
generated model responses scored by the same LLM judge, GPT-OSS-20B. The table
compares the base  model, an SFT-tuned baseline, and our GRPO-tuned
model under a consistent passage-grounded rubric-reward protocol.}

\label{tab:rubric_validation}
\begin{small}
\begin{tabular}{lc}
\toprule
\textbf{Method} & \textbf{Held-out Rubric Reward} \\
\midrule
Base & $26.1\%$ \\
SFT tuned & $41.8\%$ \\
\textbf{Ours (GRPO)} & $\mathbf{71.7\%}$ \\
\bottomrule
\end{tabular}
\end{small}
\end{table}
\begin{table*}[!ht]
\centering
\caption{\textbf{Transfer to reasoning benchmarks.}
Numbers are accuracy for the base model and the best GRPO checkpoint
selected by held-out rubric reward. None of these benchmarks is derived from the training corpus.}
\label{tab:transfer}
\begin{small}
\begin{tabular}{lccccc}
\toprule
\textbf{Method} & \textbf{GSM8K} & \textbf{MATH} & \textbf{GPQA Main} & \textbf{GPQA Diamond} & \textbf{Avg.\ $\Delta$} \\
\midrule
Base & $84.53$ & $50.06$  & $25.22$ & $24.24$ & -- \\
\textbf{Ours} & $\mathbf{85.44}$ & $\mathbf{52.88}$  & $\mathbf{33.93}$ & $\mathbf{32.32}$ & $\mathbf{+5.13}$ \\
\bottomrule
\end{tabular}
\end{small}
\end{table*}

\begin{figure*}[!ht]
\centering
\includegraphics[width=.9\linewidth]{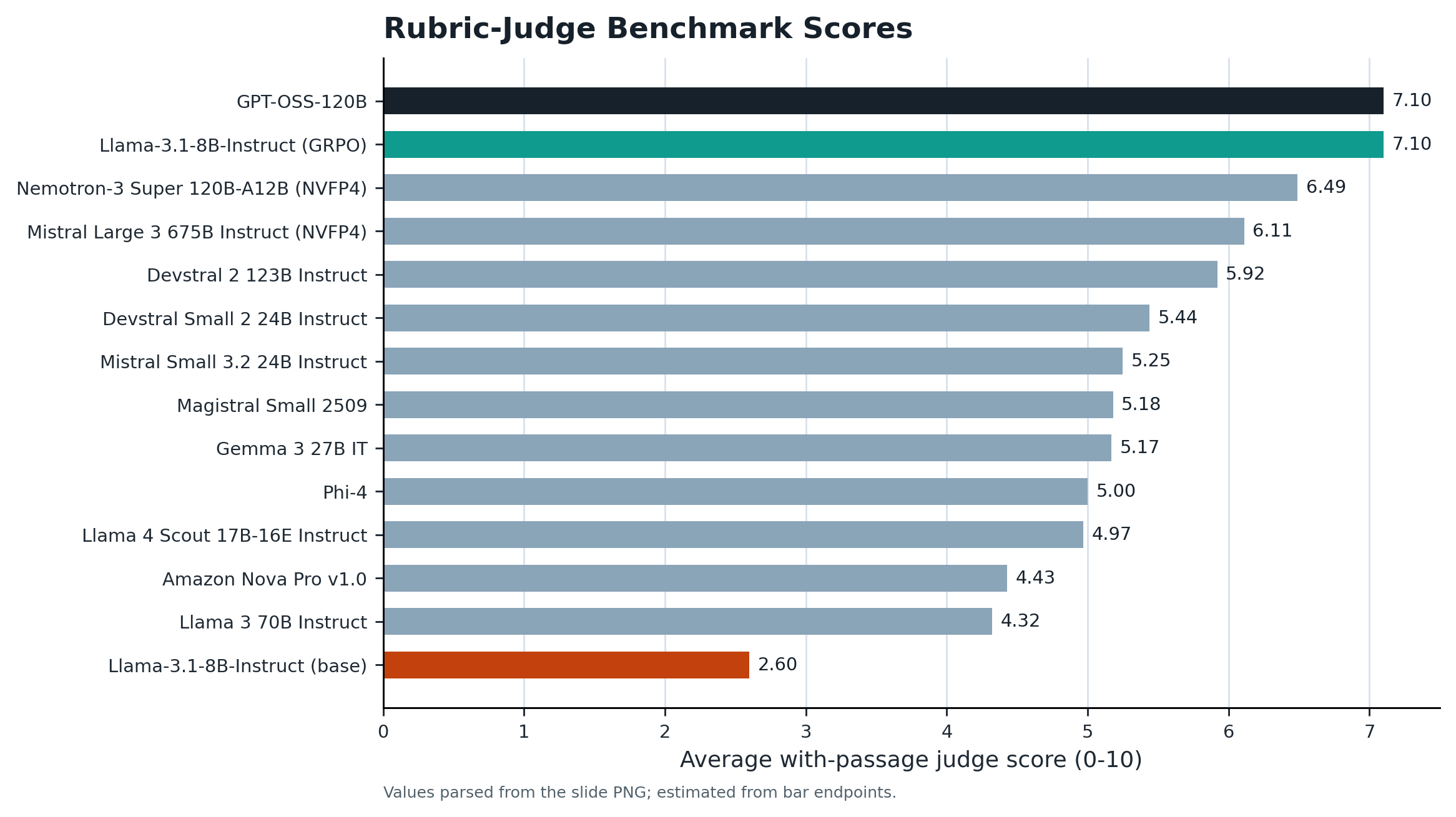}
\caption{\textbf{Rubric-judge benchmark comparison.}
In a passage-grounded rubric-judged evaluation, the RL-tuned 8B policy
substantially improves over the base 8B instruction model and approaches
the displayed GPT-OSS-120B score under the same rubric-judge protocol.}
\label{fig:judge_benchmark}
\end{figure*}
Table~\ref{tab:rubric_validation} reports an in-domain held-out rubric
evaluation under a consistent evaluation protocol. For each model, we generate
responses to the same held-out questions and score them with the same LLM judge,
GPT-OSS-20B, using the corresponding held-out rubrics and grounding passages.
Thus, the base  model, the SFT-tuned model, and the GRPO-tuned model
are compared directly under the same passage-grounded rubric-reward protocol.

The SFT model is instruction-tuned on question--answer pairs derived from the
OSTI dataset; however, its reported performance is computed exclusively on the
held-out question--rubric test split. The results show that supervised
fine-tuning improves substantially over the base  model, increasing
held-out rubric reward from $26.1\%$ to $41.8\%$. GRPO further improves the
policy from the same base initialization, achieving a held-out rubric reward of
$71.7\%$.

\Cref{fig:judge_benchmark} provides a complementary in-domain comparison under
the same passage-grounded rubric-judge protocol. The GRPO-tuned Llama-3.1-8B
policy improves from the base model's roughly $2.6/10$ judge score to roughly
$7.1/10$, approaching the displayed GPT-OSS-120B score and surpassing several
larger comparison models under the same evaluation protocol.

\subsection{Out-of-Distribution Transfer}

Table~\ref{tab:transfer} shows transfer to disjoint benchmarks. The
GRPO-tuned policy improves on all four benchmarks. The GPQA gains ($+8.71$ Main, $+8.08$ Diamond) are larger
than the gains on the math benchmarks, suggesting that rubric-grounded
training may be especially helpful for scientific reasoning tasks where
answers must coordinate facts, assumptions, and evidence. Since none of these benchmarks is derived from the training corpus, the
result is more consistent with transferable reasoning improvement than
direct document recall. The smaller GSM8K and MATH gains are also informative:
our generated questions are mostly technical question-answering tasks, not deliberately constructed quantitative-problem or code-generation tasks.
This likely makes the learned reward geometry better aligned with GPQA
than with benchmarks that require explicit calculation routines.

\subsection{Training Dynamics}

\begin{figure}[!ht]
\centering
\includegraphics[width=.46\textwidth]{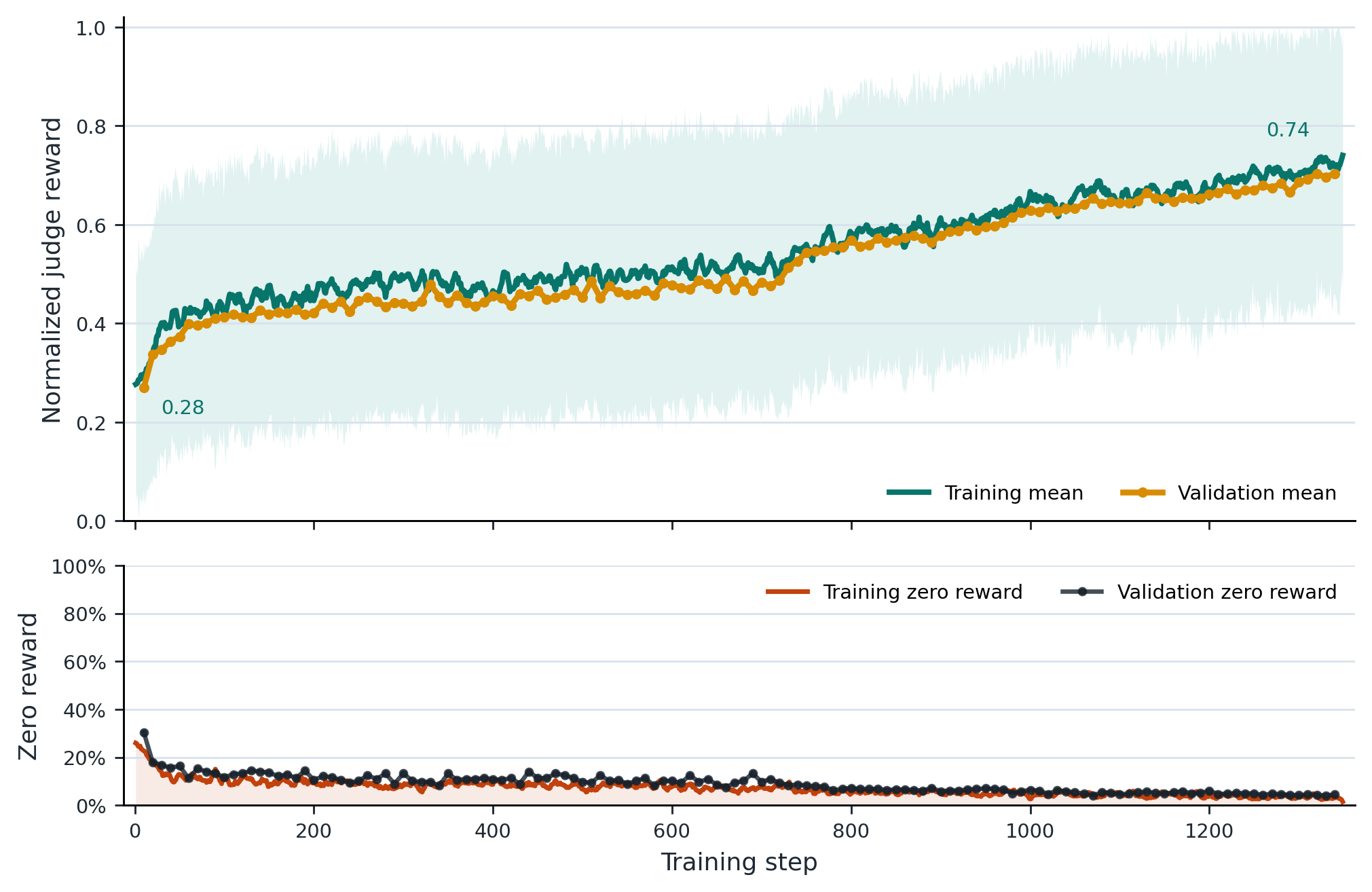}
\caption{\textbf{Reward dynamics.}
Training and validation  rewards increase over the GRPO run while
zero-reward judge outcomes decrease. The held-out trajectory supports
checkpoint selection by held-out reward and argues against improvements
being confined to sampled training prompts.}
\label{fig:learning_curves}
\end{figure}

Validation reward tracks training reward throughout the run
(\Cref{fig:learning_curves}); zero-reward judge outcomes decrease
monotonically. The validation--training gap stays bounded, which argues against
improvements being confined to sampled training prompts.

\subsection{Judge Interface Safeguards}
\label{sec:exp:reliability}

The reward signal comes from an LLM, so a reasonable  first
question is whether the policy learned to answer well or to game the
judge. We do not yet have independent human-agreement or cross-judge
reliability measurements, so we avoid treating judge reliability as a
settled empirical result. Instead, the current implementation uses four
interface safeguards.

\textbf{Low-temperature judging.}
The judge is queried at $\tau_{\mathcal{J}}=0.1$ to reduce stochastic
variation in reward assignments.

\textbf{Structured parsing.}
The judge must return schema-conforming criterion scores, a total score,
and a maximum possible score. Invalid outputs are assigned conservative
zero reward rather than repaired optimistically.

\textbf{Document-grounded scoring.}
The judge sees the source passage and rubric, while the policy sees only
the question. This makes the judge a document-grounded verifier and
prevents the policy from using retrieval context at evaluation time.

\textbf{Held-out validation.}
The same judge interface is applied to held-out rubric examples during
training. The validation trajectory in \Cref{fig:learning_curves}
therefore serves as the main check that reward increases are not confined
to the sampled training prompts.

\subsection{Analysis: What Generalizes?}
\label{sec:analysis}

The largest external gains are on GPQA Main and GPQA Diamond. This is the
most natural transfer target for the training signal: the reward asks the
policy to satisfy document-grounded scientific rubrics, and GPQA also
requires careful use of scientific concepts under uncertainty.

GSM8K and MATH improve, but by less than GPQA. This suggests that the
rubric reward is not merely teaching benchmark-specific shortcuts; rather,
it appears to improve general problem-solving behaviors that are most
visible when the downstream task resembles technical assessment.
A direct test of the reward-design hypothesis would generate document-grounded
quantitative tasks and include rubric criteria for equation setup,
calculation, units, numerical consistency, and final-answer verification.
The analogous code setting would synthesize implementation or debugging
tasks from technical documents and score criteria such as API use,
algorithmic correctness, edge cases, and test behavior.

\section{Discussion}
\label{sec:discussion}

Rubric-grounded RL is not specific to documents or to scientific
question answering. The framework requires only a checkable rubric and
an instance-level grounding the judge can use; many practical alignment
problems already have these (clinical guidelines, code-style guides,
legal frameworks, marking schemes). The key methodological move is
exposing the structure of evaluation to the optimizer rather than
collapsing it to a scalar.

The framework moves the alignment problem partly into the judge. Noisy or
inconsistent criterion scores become noisy advantages, especially when
the sampled responses for a prompt have nearly identical rewards. Low-temperature decoding, structured output, parser conservatism, reward
clipping, and held-out evaluation are therefore part of the method
rather than incidental implementation details.
Rubric grounding is most useful when (i) quality decomposes naturally
into checkable criteria, (ii) per-instance grounding is cheap to provide,
and (iii) scalar preference data is hard or expensive to collect. It is
\emph{not} the right tool when the target behavior is genuinely
holistic, when criteria are difficult to specify without domain experts,
or when the cost of grouped judge calls dominates other training cost
considerations.

The present data-generation pipeline mostly produces general
document-grounded technical questions. A natural next step is to sample
documents selectively for latent math, data-analysis, or programming
content and instruct the rubric generator to create tasks whose scoring
criteria explicitly reward quantitative setup, symbolic or numerical
calculation, unit checks, code correctness, and executable tests. This
would test whether the same rubric-grounded RL recipe can move beyond
general scientific reasoning gains and produce larger improvements on
math and code benchmarks.
\section{Limitations}
\label{sec:limitations}

The empirical study is at one model scale (8B), one corpus family
(OSTI-derived scientific and technical documents), one judge family, and
one GRPO training run. The rubric pipeline can inherit
biases of the synthesizer LLM and of the OSTI-derived source corpus.
Single-turn answering is the only setting evaluated; extending to
multi-turn interactive tasks is straightforward in the framework but not
validated here. We do not report confidence intervals or variance across
random seeds; the external benchmark improvements should therefore be
interpreted as single-run evidence rather than a stable estimate of
expected gain. Because the same judge family is used during training and
validation, held-out rubric reward does not by itself rule out
judge-specific reward hacking. Independent human-agreement or
cross-judge reliability measurements remain important directions for
future evaluation.

\section{Conclusion}
\label{sec:conclusion}

We formalized rubric-grounded RL: reinforcement learning from weighted,
judge-scored criteria grounded in instance-specific evidence unavailable
to the policy at rollout time. Instantiated on rubrics derived from
roughly 100,000 OSTI-derived scientific and technical documents, the
framework produced an 8B policy that improves held-out rubric reward and
yields gains on four reasoning benchmarks not derived from the training
corpus. The current evidence is narrower than a full ablation study: it
shows that one reported rubric-grounded GRPO run improves over the base
model under our evaluation protocol, not that every component has been
causally isolated. Even with that caveat, the result suggests a promising
post-training recipe for domains where quality is naturally rubric-shaped.

\bibliography{references}

@article{christiano2017deep,
  title={Deep Reinforcement Learning from Human Preferences},
  author={Christiano, Paul F. and Leike, Jan and Brown, Tom B. and Martic, Miljan and Legg, Shane and Amodei, Dario},
  journal={Advances in Neural Information Processing Systems},
  volume={30},
  year={2017}
}

@article{ouyang2022training,
  title={Training Language Models to Follow Instructions with Human Feedback},
  author={Ouyang, Long and Wu, Jeffrey and Jiang, Xu and Almeida, Diogo and Wainwright, Carroll L. and Mishkin, Pamela and Zhang, Chong and Agarwal, Sandhini and Slama, Katarina and Ray, Alex and others},
  journal={Advances in Neural Information Processing Systems},
  volume={35},
  pages={27730--27744},
  year={2022}
}

@article{stiennon2020learning,
  title={Learning to Summarize with Human Feedback},
  author={Stiennon, Nisan and Ouyang, Long and Wu, Jeffrey and Ziegler, Daniel M. and Lowe, Ryan and Voss, Chelsea and Radford, Alec and Amodei, Dario and Christiano, Paul F.},
  journal={Advances in Neural Information Processing Systems},
  volume={33},
  pages={3008--3021},
  year={2020}
}

@article{schulman2017proximal,
  title={Proximal Policy Optimization Algorithms},
  author={Schulman, John and Wolski, Filip and Dhariwal, Prafulla and Radford, Alec and Klimov, Oleg},
  journal={arXiv preprint arXiv:1707.06347},
  year={2017}
}

@article{schulman2020kl,
  title={Approximating {KL} Divergence},
  author={Schulman, John},
  journal={Blog post},
  year={2020}
}

@article{rafailov2023direct,
  title={Direct Preference Optimization: Your Language Model is Secretly a Reward Model},
  author={Rafailov, Rafael and Sharma, Archit and Mitchell, Eric and Ermon, Stefano and Manning, Christopher D. and Finn, Chelsea},
  journal={Advances in Neural Information Processing Systems},
  volume={36},
  year={2023}
}

@article{bai2022constitutional,
  title={Constitutional {AI}: Harmlessness from {AI} Feedback},
  author={Bai, Yuntao and Kadavath, Saurav and Kundu, Sandipan and Askell, Amanda and Kernion, Jackson and Jones, Andy and Chen, Anna and Goldie, Anna and Mirhoseini, Azalia and McKinnon, Cameron and others},
  journal={arXiv preprint arXiv:2212.08073},
  year={2022}
}

@article{lee2023rlaif,
  title={{RLAIF}: Scaling Reinforcement Learning from Human Feedback with {AI} Feedback},
  author={Lee, Harrison and Phatale, Samrat and Mansoor, Hassan and Lu, Kellie and Mesnard, Thomas and Bishop, Colton and Carbune, Victor and Rastogi, Abhinav},
  journal={arXiv preprint arXiv:2309.00267},
  year={2023}
}

@article{zheng2023judging,
  title={Judging {LLM}-as-a-Judge with {MT-Bench} and Chatbot Arena},
  author={Zheng, Lianmin and Chiang, Wei-Lin and Sheng, Ying and Zhuang, Siyuan and Wu, Zhanghao and Zhuang, Yonghao and Lin, Zi and Li, Zhuohan and Li, Dacheng and Xing, Eric P. and others},
  journal={Advances in Neural Information Processing Systems},
  volume={36},
  year={2023}
}

@inproceedings{kim2024prometheus,
  title={Prometheus: Inducing Fine-Grained Evaluation Capability in Language Models},
  author={Kim, Seungone and Shin, Jamin and Cho, Yejin and Jang, Joel and Longpre, Shayne and Lee, Hwaran and Yun, Sangdoo and Shin, Seongjin and Kim, Sungdong and Thorne, James and Seo, Minjoon},
  booktitle={International Conference on Learning Representations},
  year={2024}
}

@article{kim2024prometheus2,
  title={Prometheus 2: An Open Source Language Model Specialized in Evaluating Other Language Models},
  author={Kim, Seungone and Suk, Juyoung and Longpre, Shayne and Lin, Bill Yuchen and Shin, Jamin and Welleck, Sean and Neubig, Graham and Lee, Moontae and Lee, Kyungjae and Seo, Minjoon},
  journal={arXiv preprint arXiv:2405.01535},
  year={2024}
}

@inproceedings{ye2024flask,
  title={{FLASK}: Fine-Grained Language Model Evaluation Based on Alignment Skill Sets},
  author={Ye, Seonghyeon and Kim, Doyoung and Kim, Sungdong and Hwang, Hyeonbin and Kim, Seungone and Jo, Yongrae and Thorne, James and Kim, Juho and Seo, Minjoon},
  booktitle={International Conference on Learning Representations},
  year={2024}
}

@article{MATH,
  title={Measuring Mathematical Problem Solving With the MATH Dataset},
  author={Dan Hendrycks and Collin Burns and Saurav Kadavath and Akul Arora and Steven Basart and Eric Tang and Dawn Song and Jacob Steinhardt},
  journal={NeurIPS},
  year={2021}
}

@misc{GSM8k,
      title={Training Verifiers to Solve Math Word Problems}, 
      author={Karl Cobbe and Vineet Kosaraju and Mohammad Bavarian and Mark Chen and Heewoo Jun and Lukasz Kaiser and Matthias Plappert and Jerry Tworek and Jacob Hilton and Reiichiro Nakano and Christopher Hesse and John Schulman},
      year={2021},
      eprint={2110.14168},
      archivePrefix={arXiv},
      primaryClass={cs.LG},
      url={https://arxiv.org/abs/2110.14168}, 
}

@inproceedings{cook2024checkeval,
  title={{CheckEval}: Robust Evaluation Framework Using Large Language Model via Checklist},
  author={Lee, Yukyung and Kim, Joonghoon and Kim, Jaehee and Cho, Hyowon and Kang, Pilsung},
  booktitle={HEAL Workshop at CHI},
  year={2024}
}

@article{gunjal2025rubrics,
  title={Rubrics as Rewards: Reinforcement Learning Beyond Verifiable Domains},
  author={Gunjal, Anisha and Wang, Anthony and Lau, Elaine and Nath, Vaskar and He, Yunzhong and Liu, Bing and Hendryx, Sean},
  journal={arXiv preprint arXiv:2507.17746},
  year={2025}
}

@article{huang2025rubricanchors,
  title={Reinforcement Learning with Rubric Anchors},
  author={Huang, Zenan and Zhuang, Yihong and Lu, Guoshan and Qin, Zeyu and Xu, Haokai and Zhao, Tianyu and Peng, Ru and Hu, Jiaqi and Shen, Zhanming and Hu, Xiaomeng and others},
  journal={arXiv preprint arXiv:2508.12790},
  year={2025}
}

@article{zhou2025rubricscaffolded,
  title={Breaking the Exploration Bottleneck: Rubric-Scaffolded Reinforcement Learning for General {LLM} Reasoning},
  author={Zhou, Yang and Li, Sunzhu and Liu, Shunyu and Fang, Wenkai and Zhang, Kongcheng and Zhao, Jiale and Yang, Jingwen and Zhou, Yihe and Lv, Jianwei and Zheng, Tongya and others},
  journal={arXiv preprint arXiv:2508.16949},
  year={2025}
}

@article{lightman2023lets,
  title={Let's Verify Step by Step},
  author={Lightman, Hunter and Kosaraju, Vineet and Burda, Yura and Edwards, Harri and Baker, Bowen and Lee, Teddy and Leike, Jan and Schulman, John and Sutskever, Ilya and Cobbe, Karl},
  journal={arXiv preprint arXiv:2305.20050},
  year={2023}
}

@inproceedings{wang2024mathshepherd,
  title={{Math-Shepherd}: Verify and Reinforce {LLMs} Step-by-Step without Human Annotations},
  author={Wang, Peiyi and Li, Lei and Shao, Zhihong and Xu, Runxin and Dai, Damai and Li, Yifei and Chen, Deli and Wu, Yu and Sui, Zhifang},
  booktitle={Proceedings of the 62nd Annual Meeting of the Association for Computational Linguistics},
  pages={9426--9439},
  year={2024}
}

@article{uesato2022solving,
  title={Solving Math Word Problems with Process- and Outcome-Based Feedback},
  author={Uesato, Jonathan and Kushman, Nate and Kumar, Ramana and Song, H. Francis and Siegel, Noah Y. and Wang, Lisa and Creswell, Antonia and Irving, Geoffrey and Higgins, Irina},
  journal={arXiv preprint arXiv:2211.14275},
  year={2022}
}

@article{shao2024deepseekmath,
  title={{DeepSeekMath}: Pushing the Limits of Mathematical Reasoning in Open Language Models},
  author={Shao, Zhihong and Wang, Peiyi and Zhu, Qihao and Xu, Runxin and Song, Junxiao and Zhang, Mingchuan and Li, Y. K. and Wu, Y. and Guo, Daya},
  journal={arXiv preprint arXiv:2402.03300},
  year={2024}
}

@article{yu2025dapo,
  title={{DAPO}: An Open-Source {LLM} Reinforcement Learning System at Scale},
  author={Yu, Qiying and Zhang, Zheng and Zhu, Ruofei and Yuan, Yufeng and Zuo, Xiaochen and Yue, Yu and Dai, Weinan and Fan, Tiantian and Liu, Gaohong and Liu, Lingjun and others},
  journal={arXiv preprint arXiv:2503.14476},
  year={2025}
}

@article{zheng2025gspo,
  title={Group Sequence Policy Optimization},
  author={Zheng, Chujie and Liu, Shixuan and Li, Mingze and Chen, Xiong-Hui and Yu, Bowen and Gao, Chang and Dang, Kai and Liu, Yuqiong and Men, Rui and Yang, An and Zhou, Jingren and Lin, Junyang},
  journal={arXiv preprint arXiv:2507.18071},
  year={2025}
}

@article{dubey2024llama3,
  title={The {Llama} 3 Herd of Models},
  author={Dubey, Abhimanyu and Jauhri, Abhinav and Pandey, Abhinav and Kadian, Abhishek and Al-Dahle, Ahmad and Letman, Aiesha and Mathur, Akhil and Schelten, Alan and Yang, Amy and Fan, Angela and others},
  journal={arXiv preprint arXiv:2407.21783},
  year={2024}
}

@article{cobbe2021training,
  title={Training Verifiers to Solve Math Word Problems},
  author={Cobbe, Karl and Kosaraju, Vineet and Bavarian, Mohammad and Chen, Mark and Jun, Heewoo and Kaiser, Lukasz and Plappert, Matthias and Tworek, Jerry and Hilton, Jacob and Nakano, Reiichiro and Hesse, Christopher and Schulman, John},
  journal={arXiv preprint arXiv:2110.14168},
  year={2021}
}

@article{hendrycks2021measuring,
  title={Measuring Mathematical Problem Solving with the {MATH} Dataset},
  author={Hendrycks, Dan and Burns, Collin and Kadavath, Saurav and Arora, Akul and Basart, Steven and Tang, Eric and Song, Dawn and Steinhardt, Jacob},
  journal={arXiv preprint arXiv:2103.03874},
  year={2021}
}

@article{rein2023gpqa,
  title={{GPQA}: A Graduate-Level Google-Proof {Q\&A} Benchmark},
  author={Rein, David and Hou, Betty Li and Stickland, Asa Cooper and Petty, Jackson and Pang, Richard Yuanzhe and Dirani, Julien and Michael, Julian and Bowman, Samuel R.},
  journal={arXiv preprint arXiv:2311.12022},
  year={2023}
}
\bibliographystyle{icml2026}

\newpage
\appendix

\section{Full Algorithm}
\label{app:algorithm}

\begin{algorithm}[t]
\caption{Rubric-Grounded GRPO}
\label{alg:grpo}
\begin{algorithmic}[1]
\REQUIRE Policy $\pitheta$, reference $\piref$, judge $\mathcal{J}$,
  dataset $\dataset$, group size $G$, clip $\epsclip$, stabilizer
  $\delta$, KL coefficient $\beta$, learning rate $\eta$, max gradient
  norm $\gamma$
\STATE Initialize $\pigen \leftarrow \pitheta$
\FOR{epoch $=1,\ldots,E$}
  \FOR{batch $\{(\prompt_i, p_i, \rubric_i)\}_{i=1}^B \sim \dataset$}
    \STATE Synchronize $\pigen \leftarrow \pitheta$
    \FOR{$i=1,\ldots,B$}
      \STATE Sample $\{\response_i^{(g)}\}_{g=1}^G \sim
             \pigen(\cdot\mid\prompt_i)$; record $\log\pigen$.
    \ENDFOR
    \FOR{$i=1,\ldots,B;\;g=1,\ldots,G$ \textbf{(parallel)}}
      \STATE $R_i^{(g)} \leftarrow
             \mathcal{J}(\prompt_i, p_i, \response_i^{(g)},
             \rubric_i) / W_i$
    \ENDFOR
    \FOR{$i=1,\ldots,B;\;g=1,\ldots,G$}
      \STATE $b_i^{(g)} \leftarrow
             \tfrac{1}{G-1}\sum_{g'\neq g} R_i^{(g')}$;
             $\sigma_i \leftarrow \mathrm{std}(R_i)$
      \STATE $A_i^{(g)} \leftarrow (R_i^{(g)} - b_i^{(g)})/
             (\sigma_i+\delta)$ if $\sigma_i>0$, else $0$
    \ENDFOR
    \STATE Compute $\log\pitheta, \log\piref$ via forward passes
    \FOR{each response token $(i,g,t)$}
      \STATE $r_t \leftarrow \exp(\log\pitheta - \log\pigen)$
      \STATE $\ell_t \leftarrow -\min(r_t A_t,
             \mathrm{clip}(r_t, 1{-}\epsclip, 1{+}\epsclip) A_t)$
      \STATE $\widehat{\KL}_t \leftarrow
             \exp(\log\piref - \log\pitheta) - 1
             - (\log\piref - \log\pitheta)$
    \ENDFOR
    \STATE $\loss \leftarrow \mathrm{mean}(\ell_t + \beta\widehat{\KL}_t)$
    \STATE $\theta \leftarrow \theta - \eta\,
           \mathrm{clip\_grad}(\nabla_\theta\loss, \gamma)$
  \ENDFOR
\ENDFOR
\end{algorithmic}
\end{algorithm}

\section{Estimator Notes}
\label{app:estimators}

\paragraph{LOO baseline.}
For i.i.d.\ rewards $R_1,\ldots,R_G$ with mean $\mu$ and variance
$\sigma^2$, the leave-one-out baseline
$b^{(g)}=\tfrac{1}{G-1}\sum_{g'\neq g}R_{g'}$ satisfies
$\E[b^{(g)}]=\mu$, $\mathrm{Cov}(R_g,b^{(g)})=0$, and
$\mathrm{Var}(b^{(g)})=\sigma^2/(G-1)$.

\paragraph{$k_3$ KL approximation.}
The approximation $\widehat{\KL}=e^u-1-u$ is nonnegative by convexity
($e^u\geq 1+u$), equals zero iff $u=0$, and has Taylor expansion
$\widehat{\KL}=u^2/2+O(u^3)$.

\paragraph{Judge-noise heuristic.}
Let the normalized reward be
$r=\sum_j w_j s_j/W$, where $W=\sum_j w_j$. If criterion score noise is
zero-mean, independent, and has common variance $\tau^2$, then the
aggregate noise variance is
\begin{equation}
  \mathrm{Var}(r)
  = \tau^2\frac{\sum_j w_j^2}{W^2}
  = \frac{\tau^2}{M_{\mathrm{eff}}},
  \quad
  M_{\mathrm{eff}}=\frac{W^2}{\sum_j w_j^2}.
\end{equation}
Thus equally weighted criteria reduce independent criterion noise by a
factor of $M$, and uneven weights reduce it by the effective number of
criteria $M_{\mathrm{eff}}$. This motivates multi-criterion rubrics but
is not used as a formal reliability guarantee in the main paper.

\section{Variants Supported but Not Used in the Main Run}
\label{app:variants}

\paragraph{Importance sampling.}
For multi-step updates per rollout, token-level IS uses
$w_t=\exp(\log\pitheta-\log\pigen)$. Sequence-level IS (GSPO) uses the
geometric mean of per-token ratios. Truncated variants (TIS, ICEPOP,
Seq-Mask-TIS) cap or mask weights outside a configured band.

\paragraph{Dual clipping.}
For $A_t<0$, $\ell_t \leftarrow \max(\ell_t^{\mathrm{clip}}, -c A_t)$
with $c>1$ (typically $3$).

\paragraph{Dynamic sampling (DAPO).}
Filter prompts with $\sigma_i=0$ and re-sample until the batch contains
the required number of informative prompts.

\paragraph{Reward shaping.}
A length penalty discourages near-truncation responses; a truncation
penalty multiplies reward by $\alpha_{\mathrm{stop}}\in[0,1)$ when the
response lacks a stop token. Both are disabled in main runs.

\section{Judge Prompt Template}
\label{app:judge_prompt}

\begin{verbatim}
SYSTEM: You are a strict, objective
academic evaluator. Score the RESPONSE
against each evaluation criterion using
the provided scoring guide, required
elements, and expected keywords. Return
ONLY a valid JSON object.

USER:
SOURCE PASSAGE: [g, truncated]
QUESTION: [x]
RESPONSE: [y]
CRITERIA (total weight W):
  1. name, w_1, eta_1,
     required E_1, guide, kappa_1, nu_1
  ...
Return:
{ "scores": {"c_1": s_1, ...},
  "total": <sum>, "max_total": <W>,
  "reasoning": "<brief>" }
\end{verbatim}

\section{Data Schema and Pipeline Configuration}
\label{app:data_pipeline}

\begin{verbatim}
{
  "question": <string>,
  "passage":  <string>,
  "criteria": [
    { "id": <string>, "weight": <float>,
      "name": <string>,
      "description": <string>,
      "required_elements": [<string>],
      "scoring_guide": <string>,
      "verification_method": <string>,
      "expected_keywords": [<string>],
      "expected_concepts": [<string>] },
    ...
  ],
  "question_rationale": <string>,
  "document_analysis": {
    "genre": <string>, "contribution": <string>,
    "concepts": [<string>], "depth": <string>,
    "reasoning_mode": <string>
  },
  "doc_hash": <string>
}
\end{verbatim}

\end{document}